\documentclass[runningheads]{llncs}
\pdfoutput=1

\usepackage{mathptmx}
\usepackage{comment}

\title{Planning and Scheduling in Hybrid Domains Using Answer Set Programming }
\author{Sandeep Chintabathina} 
\institute{University of Arkansas at Pine Bluff, Pine Bluff AR 71601, USA \\ \email{chintabathinas@uapb.edu} }
\titlerunning{Planning and Scheduling in Hybrid Domains Using Answer Set Programming}
\authorrunning{S.~Chintabathina} 

\begin{document}

\setcounter{page}{67}

\maketitle




\begin{abstract}
In this paper we present an Action Language-Answer Set Programming based approach to solving planning and scheduling problems in \emph{hybrid} domains - domains that exhibit both discrete and continuous behavior.  We use action language H to represent the domain and then translate the resulting theory into an A-Prolog program. In this way, we reduce the problem of finding solutions to planning and scheduling problems to computing answer sets of A-Prolog programs. We cite a planning and scheduling example from the literature and show how to model it in H. 
We show how to translate the resulting H theory into an equivalent A-Prolog program. We compute the answer sets of the resulting program using a hybrid solver called EZCSP which loosely integrates a constraint solver with an answer set solver. 
The solver allows us reason about constraints over reals and compute solutions to complex planning and scheduling problems.  Results have shown that our approach can be applied to any planning and scheduling problem in hybrid domains. 

\end{abstract}




\section{Introduction}
In this paper we are interested in modeling intelligent agents capable of planning, acting and reasoning in a dynamic environment. We are primarily interested in \emph{hybrid domains} - domains that exhibit both discrete and continuous 
behavior.  Many real world situations such as a robot in a manufacturing plant,  a decision control system of a space shuttle etc. deal with both discrete change as well as continuous change.  For example, a decision control system in a space shuttle is capable of opening and closing valves of fuel tanks to supply fuel to some jets. The actions \emph{open} and \emph{close} change positions of valves. The valves will remain in a certain position as long as no \emph{open} or \emph{close} actions occurs. The change of position of valves is, therefore, discrete. However, the fuel level in the tank can change continuously with time when the valve is open or there is an incoming supply of fuel to the tank. This type of continuous change coupled with discrete change makes this domain hybrid. We are not only interested in modeling such domains but also to solve planning and scheduling problems in such domains. A simple planning and scheduling problem in the above example would be to fire a jet within 10 seconds. This would require opening and closing the appropriate valves and delivering fuel to the jet in time to achieve the goal. 

To model intelligent agents we need to provide the agent with knowledge about its environment 
and its own capabilities and goals.  There are several approaches to representing and reasoning about such knowledge. In this paper, we use action language $H$ - a high level language for representing and reasoning about actions and their effects. We chose this language mainly because it is capable of representing and reasoning about hybrid domains. A description written in H describes  a transition diagram whose states correspond
to possible physical states of the system and whose arcs are labeled by actions. A transition, 
$\langle \sigma ,a,\sigma ^{\prime} \rangle$, of a diagram denotes that action $a$ is possible in
state $\sigma$ and that after the execution of $a$ the system may move to state $\sigma^{\prime}$. 
The diagram consists of all possible trajectories of the system.  


Language H \cite{cw12} was developed by extending the signature of  action language ${\cal AL}$ \cite{bg00} with a collection of numbers for representing time, a collection of functions of time, and \emph{fluents} \footnote{functions whose values depend on a state and may change as a result of actions} with non-boolean values including fluents defined by functions of time (\emph{process fluents}).  It also allows  \emph{triggers} which specifies conditions under which actions are triggered. The semantics of ${\cal AL}$ is based on the McCain-Turner equation \cite{mt95}. A  slightly modified version of this equation defines the semantics of H.  
Thus, both languages are based on the same underlying intuition. However, in ${\cal AL}$ a state of the transition diagram is a snapshot of the world, whereas in H, a state is a snapshot of the world over a time interval.  Process fluents are defined over this time interval. 
Prior to H,  logic-based formalisms such as
Situation Calculus  \cite{r01} and Event Calculus  \cite{sha97} were extended to reason about process fluents. 
Both approaches demonstrate via examples how their approach can be used for reasoning
about process fluents. However, in both approaches,  it is difficult to express causal relations between fluents.

To solve planning and scheduling problems in hybrid domains, we use an action language-logic programming approach. We begin by representing the domain in H. To this extent, we  
 come up with a theory of H (also called \emph{action description}) to  describe actions and their effects. 
 We implement this theory by translating it into an 
\emph{Answer Set Prolog} (A-Prolog) program, a class of logic programs under \emph{answer set} semantics  \cite{gl88,gl91}, and computing answer sets (models) of the resulting program.
In this way, we reduce the problem of finding solutions to the planning and scheduling problem to computing answer sets of the A-Prolog program. In \cite{chi10} we show that there is a one-to-one correspondence between models of A-Prolog programs and models of our specification. 

The paper is organized as follows. In section 2, we revisit the syntax and semantics of H as described in \cite{cw12}. 
In section 3, we cite a planning and scheduling example from \cite{pw94} and come up with a theory of H to model it. 
In section 4, we present the main contribution of this paper which is the translation of the resulting H theory into an A-Prolog program. The example from \cite{pw94} was run using ZENO, a least  commitment planner that supports continuous change. After ZENO, researchers  have come up with several efficient solvers. One significant contribution is the Planning Domain Description Language (\emph{PDDL}) which was exclusively developed for planning purposes.  A variant of  this language, $PDDL+$ \cite{foxlong06} is capable of modeling continuous change through the use of processes and events.
 Like H, the semantics of this language is defined using a labeled transition diagram. The language has been shown to represent a number of complex time dependent effects. However, one major limitation of the language is that it does not support \emph{derived predicates} - predicates defined in terms of other predicates for eg. $above(x,y)$ from the blocks world domain.  There are no such limitations in H.

\section{Preliminaries}
In this section we revisit the syntax and semantics of H as described in \cite{cw12}.
\subsection{Syntax}
By \emph{sort} we mean a non-empty countable collection of strings in some 
fixed alphabet. A \emph{sorted signature} $\Sigma$ is a collection of sorts 
and function symbols. 
A \emph{process signature} is a sorted signature with special sorts
\emph{time}, \emph{action}, and \emph{process}.
Sort \emph{time} is normally identified with one of the standard numerical
sorts with the exception that it contains an ordinal $\omega$ such that
for any $x \in time \setminus \{\omega\}$, $\omega >x$.  No operations are
defined over $\omega$. If time is discrete, elements of $time \setminus \{\omega\}$ may be viewed as
non-negative integers, otherwise they can be interpreted as either rational numbers, 
constructive real numbers, etc. 

\noindent
Sort \emph{process} contains strings of the form $\lambda T. f(T)$ where $T$ is a variable
ranging over \emph{time} and $f(T)$ is a mathematical expression (possibly) containing $T$.
A string $\lambda T.f(T)$ represents a function defined over \emph{time}. The $\lambda$ is said to bind $T$ in $f(T)$. 
If  the expression, $f(T)$, does not contain any variables for time then $\lambda T.f(T)$ is said to denote a constant function. For e.g. $\lambda T. 0$ denotes the constant function 0. For simplicity we assume that all functions from \emph{process} have the same range denoted by the sort $range(process)$. 
An example of a function from sort \emph{process} is $\lambda T. h - (g/2)*(T-t)^2$ which defines the
height in meters, at time T, of a freely falling object dropped from a height $h$, $T-t$ seconds before. The symbol $g$ denotes the Earth's average gravitational acceleration which is equal to $9.8$ $meters/sec^{2}$. 
 

\noindent
Sort \emph{action} is divided into subsorts \emph{agent} and \emph{exogenous}. Members of \emph{agent} denote
(unit) actions performed by an agent and members of \emph{exogenous} denote non-agent actions. 
Both agent and exogenous actions will be referred to as actions. A \emph{compound action} is a set of unit actions performed at the same time.


\noindent
The collection of function symbols includes names for fluents and standard numerical functions.
Each fluent name is associated with an \emph{arity} - a number indicating the number of arguments. 
Intuitively, \emph{fluents} are properties that may change as a result of actions.
For example, the \emph{height} of a brick held at a certain position above the ground could
change when it is dropped. 
Every process signature contains {\it reserved fluents} {\bf start} and {\bf end} of sort $time$.




 


\noindent
A \emph{term of sort} $s$ is defined as follows:
\begin{enumerate}
{
\item
A string $y \in s$ is a term of sort $s$;
\item
If $t_1,\dots,t_n$ are terms of sorts $s_1,\dots,s_n$ respectively and  $f: s_1 \times \dots \times s_n \rightarrow s$ is a function symbol then $f(t_1,\dots,t_n)$ is  a term of sort $s$.


}
\end{enumerate}

\noindent
Notice that if $f(\bar{x})$ is a term of sort \emph{process} and $t$ is a term of sort \emph{time}
then $f(\bar{x})(t)$ is a term of sort $range(process)$.
For example, to represent the height of brick $b$ we can introduce a fluent $height(b)$ of sort $process$.
Then by $height(b)(10)$ we denote the height of $b$ at time 10.  Similarly, $\lambda T. 200- (g/2)*(T-5)^2(10)$  denotes the value of the function at $T=10$ which is equal to 77.5.





\noindent
An \emph{atom} of $\Sigma$ is a statement of the form $t = y$ where $t$ is a term of some sort $s$ and $y \in s$. 
Examples of atoms are $end=10$,  $2+3=5$ etc. 
If $t$ is a term built from fluent symbols then such an atom is called a \emph{fluent atom}.
Examples of fluent atoms are $height(b)=\lambda T. 100- (g/2)*(T-5)^2$, $height(b)(5)=100$, $end=10$, etc.

\noindent
A \emph{literal} of $\Sigma$ is an atom $t=y$ or its negation $\neg(t=y)$ . Negation of $=$ will be often
written as $\neq$. If $t$ is a term of Boolean sort then $t=true$ ($t\neq false$) is often
written as $t$ and $t=false$ ($t\neq true$) is often written as $\neg t$.  For example, the atom $4<5= true$ will
be written as $4<5$.

\noindent
Language, H, is parameterized by a process signature $\Sigma$ with standard interpretations of numerical
functions and relations (such as $+,<,\leq,\neq,$ etc).
\begin{definition}
{\rm
An \emph{action description} of $H(\Sigma)$ is a collection of statements of the form:
\[
\begin{array}{lr}
l_0 \mbox{ if  } l_1,\dots,l_n. & (1)\\
e \mbox{ causes } l_0 \mbox{ if } l_1,\dots,l_n. & (2)\\
\mbox{impossible } e_1,\dots,e_m \mbox{ if } l_1,\dots,l_n. & (3)\\
l_1,\dots,l_n  \mbox{ triggers }  e.  & (4)
\end{array}
\]
}
\end{definition}
where $e$'s are elements of \emph{action},  
$l_0$'s are fluent atoms  and $l_1,\dots,l_n$ are literals of the signature of $H$.
$l_0$ is referred to as the \emph{head} of a statement and $l_1,\dots,l_n$ are
referred to as the \emph{body} of a statement.
A statement of the form (1) is called a \emph{state constraint}. It guarantees that
any state satisfying $l_1,\dots,l_n$ also satisfies $l_0$. A statement of the form (2) is
called a \emph{dynamic causal law} and it states that if
action $e$ were executed in a state satisfying literals $l_1,\dots,l_n$
then any successor state would satisfy $l_0$. A statement of the form  (3) is called an 
\emph{executability condition} and it states that
actions $e_1,\dots,e_m$ cannot be executed in a state satisfying  $l_1,\dots,l_n$.
If $n = 0$ then \emph{if} is dropped from statements (1), (2) and (3).
A statement of the form (4) is called a \emph{trigger} and it states that action $e$ is
triggered in any state satisfying $l_1,\dots,l_n$.

By ground instantiations of a variable of sort $s$, we mean the elements of $s$.
From the description of the syntax, the only variable that appears in the statements of H 
is the variable $T$ ranging over time. 
However, variables ranging over other sorts are allowed in the statements as long as those
statements are viewed as a shorthand for the collection of statements 
obtained by replacing each occurrence of a variables other than $T$ by its corresponding ground 
instantiations.

\subsection{Semantics}
The semantics of language H is based on a slightly modified McCain-Turner equation \cite{mt95}. 
An action description, $AD$, of H($\Sigma$) describes a transition diagram, $TD(AD)$, whose nodes correspond to possible
physical states of a system and whose arcs are labeled by actions. A transition $\langle s,a,s'\rangle$
of the diagram denotes that action $a$ is possible in $s$ and as a result of execution of $a$ the system may 
move to state $s'$. It is important to note that an action description of H can be either \emph{deterministic} (i.e. for any state-action pair there is at most one successor state \cite{bg00}) or \emph{non-deterministic} (i.e. there is a state-action pair with more than one successor state).  In this section we will give a formal definition for a state and a transition of $TD(AD)$. 
We begin with interpreting symbols of $\Sigma$.



\begin{definition}
{\rm
Given an action description $AD$ of $H(\Sigma)$, an \emph{interpretation} $I$  of $\Sigma$ is a mapping
defined as follows.
\begin{itemize}
\item for every non-process sort, $s$, and every string $y \in s$, $I$ maps $y$ into itself i.e. $y^I= y$. 
\item standard interpretation is used for the sort $process$ and other standard numerical functions and relations.
\item $I$ maps every fluent into a properly typed function.



\end{itemize}
}
\end{definition}






\noindent
Often an interpretation $I$ of $\Sigma$ is identified with a collection, $s(I)$, of atoms of the form $t=y$ such that
$t^I=y$ where $t$ and $y$ are terms of some sort. In other words, $s(I)=\{ t=y \mid t^I=y \}$. 

\noindent
Before we give the definition of a state of $TD(AD)$ let us consider the following definitions. 

\noindent
A set, $s$, of atoms is said to be \emph{consistent} if for every atom $t=y_1 \in s$, $\not\exists y_2$  such that $t=y_2 \in s$ and $y_1 \neq y_2$. 

\noindent
Let us define what it means for a literal to be true w.r.t a set of atoms of $\Sigma$.
\begin{definition}
{\rm
Given a consistent set, $L$, of atoms of $\Sigma$ 
\begin{itemize}
\item
An atom $t=y$ is \emph{true in L} (symbolically $L \models t=y$) iff $t=y \in L$.

\item
A literal $t \neq y$ is \emph{true in L} $(L \models t \neq y)$ iff  $L \models t=y_0 $ and $y \neq y_0$.
\end{itemize}
}
\end{definition}

\noindent
We will now define what it means for a set of atoms to be closed under state constraints of $AD$.

\begin{definition}
{\rm
A \emph{set L of atoms is closed under the state constraint} 
\[ l_0 \mbox { if } l_1,\dots,l_n \]
of $AD$ if,  whenever $L \models l_i$ for every $i$, $1 \leq i \leq n$,  $L \models l_0$.

}
\end{definition}

\noindent
A \emph{set L of atoms is closed under state constraints} of $AD$ if $L$ is closed under every state constraint of $AD$. 

\noindent
Next, we define what it means for a set of atoms to satisfy a trigger of $AD$.
\begin{definition}
{\rm
A \emph{set L of atoms of H satisfies a trigger}
\[ l_1,\dots,l_n \mbox{ triggers } e \]
of $AD$ iff $L \models l_i$ for every $i$ such that $1 \leq i \leq n$.
}
\end{definition}

\noindent
Intuitively, if a set of atoms satisfies a trigger it means that 
the corresponding action will take place at some time point. 
The next definition characterizes sets of atoms which ``$end$'' before any triggered action takes place.





\begin{definition}
{\rm
A \emph{set L of atoms of H is closed under triggers} of $AD$  
iff $\not\exists$ $L^{\prime}$ such that $L^{\prime}$ satisfies at least one
trigger of $AD$ and $L \setminus L^{\prime} =\{end= t_2\}$ and
$L^{\prime} \setminus L = \{end=t_1\}$ and  $t_1< t_2$.
}
\end{definition}
 


\noindent
Intuitively, a state of $TD(AD)$ can be viewed as a collection of functions of time defined over an interval. The endpoints
of the interval are implicitly defined by the reserved fluents $start$ and $end$. The domain of each function is the
set $\{ t\mid start\leq t \leq end \wedge t < \omega\}$. We say that a state is defined over an interval of the form
$[start,end]$  iff $end \neq \omega$. There is at least one arc labeled by an action leading out of such a state.
We say that a state is defined over an interval of the form $[start,end)$ iff $end =\omega$. 
There is no arc leading out of such a state. States that begin at time 0 are called \emph{initial states}. 
They define the initial conditions of a domain. Here is the formal definition of a state.

\begin{definition}
{\rm
Given an interpretation $I$ of $\Sigma$, $s(I)$ is a \emph{state} of $TD(AD)$ if each of the following holds. 
\begin{itemize}
\item
$s(I)$ is a collection of atoms of the form $t=y$ such that $t^I=y$ where $t$ and $y$ are terms of the same sort.
\item
$s(I)$ is closed under the state constraints of $AD$.
\item
If $s(I) \models start=t_1$  and $s(I)\models end=t_2$ then $t_1 \leq t_2 \wedge t_1 < \omega$. 
\item
$s(I)$ is closed under the triggers of $AD$.
\item
If $s(I) \models p= \lambda T. f(T)$ where $p$ is a fluent of sort $process$ then  
$\lambda T.f(T)$ is defined over the domain $\{ t \mid  start^I \leq t \leq end^I \wedge t < \omega \}$.
\item
If $p$ is a fluent of sort \emph{process} and $t$ is a term of sort $time$ then
$s(I) \models p(t)=x$ iff  $s(I) \models p=\lambda T.f(T)$ and $\lambda T. f(T)(t^I)=x$.

 

\end{itemize}
}
\end{definition}

\noindent
By definition of interpretation every symbol is mapped uniquely. Therefore, states of $TD(AD)$ are complete
and consistent. Whenever convenient the parameter $I$ will be dropped from $s(I)$. 
Next, we will define what is means for an action to be possible in a state.




\begin{definition}
{\rm
Action $a$ is \emph{possible} in state $s$ if for every non-empty subset
$a_0$ of $a$, there is no executability condition
\[ \mbox{impossible } a_0 \mbox{  if  }l_1,\dots, l_n. \]
of $AD$ such that $s \models l_i$ for every $i$, $1 \leq i \leq n$.
}
\end{definition}

\noindent
Given a state $s$ and action $e$ let us define what are the direct effects of executing
$e$ in $s$.

\begin{definition}
{\rm
Let $e$ be an elementary action that is possible in state $s$. By $E_s(e)$ we denote
the set of all \emph{direct effects of} $e$ w.r.t $s$.
\[ E_s(e) =  \{ l_0  \mid e \mbox{ causes } l_0 \mbox{ if } l_1,\dots,l_n  \in AD \wedge
                   s\models l_i \mbox{ for every } i, 1\leq i \leq n \}    \]
}
\end{definition}

\noindent
If $a$ is a compound action then $E_s(a) = \mathop{\bigcup}_{e \in a} E_s(e)$.




\noindent
The following definition allows us to identify sets of atoms with adjacent intervals.

\begin{definition}
{\rm
Let $x,y,$ and $z$ be elements of sort \emph{time} such that  $x \leq y \leq z \wedge y < \omega$ 
and $s$ and $s^\prime$ be sets of atoms of H. We say that $s^\prime$ \emph{follows} $s$ iff  
$s \models \{ start=x, end=y\}$ and $s^\prime \models \{start=y,end=z\}$.
}
\end{definition}

  
\noindent
Given two sets of atoms $s$ and $s^{\prime}$, the function $T_s(s^\prime)$  is defined as follows.
\[  T_s(s^{\prime})= \left \{ \begin{array}[c]{l} \{start=t_1, end=t_2 \}  \mbox{ if }s^\prime \mbox{ follows } s  \wedge  
                                 \{start=t_1, end=t_2\} \subseteq s^{\prime}.\\ 
                               \emptyset \mbox{ otherwise}. \end{array} \right.
\]
In other words, the function returns the interval of $s^\prime$ if $s^\prime$ follows $s$; otherwise it returns an empty set. 


\noindent
The consequences of a set of atoms w.r.t a set of state constraints is defined as follows.
 
\begin{definition}
{\rm
Given a set $S$ of atoms and a set $Z$ of state constraints of $AD$
the set, $Cn_Z(S)$, of \emph{consequences of $S$ under Z} is the smallest set of atoms (w.r.t
set theoretic inclusion) containing $S$ and closed under $Z$.
}
\end{definition}








\begin{definition}  
{\rm
\emph{Action $a$ is complete w.r.t a set of atoms s} if for every trigger $r \in AD$
of the form
\[ l_1,\dots,l_n \mbox{ triggers } e \]
$e\in a$ iff $s$ satisfies $r$.  
}
\end{definition}

\noindent
We know that a state contains arbitrary atoms of $\Sigma$. However, 
for the next definition we focus  only on fluent atoms including those
formed from $start$ and $end$.  All other atoms such as $2+3=5$ which is always true will be ignored. 


\begin{definition}
{\rm
A transition diagram $TD(AD)$ is a tuple $\langle \phi,\psi \rangle$ where
\begin{itemize}
\item
$\phi$ is the set of states.  
\item
$\psi$ is the set of all transitions $\langle s, a, s^{\prime}\rangle$ such that each of the following holds.
\begin{itemize}
\item $a$ is complete w.r.t $s$
\item $a$ is possible in $s$  
\item $s^\prime$ is closed under the triggers of $AD$       
\item 
\begin{equation}
\label{mct}
s^{\prime} = Cn_{Z}( E_s(a) \cup (s \cap s^{\prime} ) \cup T_s(s^{\prime}) )
\end{equation}
where $Z$ is the set of state constraints of $AD$.
\end{itemize}
\end{itemize}
The set, $E_s(a)$, consists of direct effects of $a$ while the set, $s \cap s^{\prime}$, 
consists of facts preserved by inertia. $T_s(s^{\prime})$ projects the $start$ and $end$ of $s^{\prime}$. 
The application of $Cn_{Z}$ to the union of these sets adds the indirect effects. 
From the definition, it is evident that concurrent actions such as dropping two balls from different heights at the same time can be handled in H.
}
\end{definition}

\section{A planning and scheduling example}
In this section we will visit a planning and scheduling example from \cite{pw94} 
and demonstrate how to model it in H. 
Consider a toy world in which a single plane moves passengers between cities. ÒSlow flyingÓ travels at 400 miles per hour and consumes 1 gallon of fuel every 3 miles, on average. ÒFast flyingÓ travels at 600 miles per hour and consumes 1 gallon of fuel every 2 miles. Passengers can be boarded in 30 minutes and deplaned in 20 minutes. Refueling gradually increases the fuel level to a maximum of 750 gallons, taking one hour from an empty tank. Boarding, deplaning, and refueling must all occur while the plane is on the ground. The distance between city-a and city-b is 600 miles, the distance between city-a and city-c is 1000 miles, the distance between city-b and city-c is 800 miles and the distance between city-c and city-d is 1000 miles. Suppose that Dan and Ernie are at city-c, but the empty plane and Scott are at city-a. If the plane only has 500 gallons of fuel, how can we ensure that Scott and Ernie get to city-d in less than five and a half hours? \cite{pw94}

\noindent
A solution to this problem requires reasoning about effects of (concurrent) actions in the presence of continuous time. There are properties of the domain that change continuously with time such as the fuel level of the plane, the distance covered by the plane etc. With proper planning and scheduling, the plane will make its tight schedule without running out of fuel. Hence, the problem is a planning and scheduling problem in the presence of continuous change.  
We begin with the representation of the domain in H.


\subsection{Representation in H}
We begin with the description of the signature. It consists of several objects -  persons \emph{scott, ernie,} and \emph{dan} and locations \emph{a,b,c,} and \emph{d}. We will use another location called \emph{enroute} to denote that the plane is in the air. We will use (possibly indexed)   variables $P$ and $L$ to denote persons and locations respectively. We are given the constants $distance(a,b,600)$, $distance(a,c,1000)$ and so on. We are also given the fuel consumption rates for different speeds.  For example, when flying at 400 mph the mileage is 3 miles per gallon. We can encode this information in the form of constants $fc(400)$ and
$fc(600)$ which denote 3 miles per gallon and 2 miles per gallon respectively. For modeling purposes we will use minutes as our unit of time.

\noindent
We know that there are several durative actions in the domain, for example, boarding which takes 30 minutes. We will use the approach used by Reiter \cite{r01} to model durative actions.  We introduce the action $start\_boarding(P,L)$ which denotes that person $P$ is boarding at location $L$ and $end\_boarding(P,L)$  which denotes that person $P$ has finished boarding at location $L$. So instead of one action we have two actions $start\_boarding(P,L)$ and $end\_boarding(P,L)$ which make the fluent $boarding(P,L)$ \emph{true} and \emph{false} respectively.  In our model,  the \emph{end\_boarding(P,L)} action will be triggered 30 minutes from the time $start\_boarding(P,L)$ is executed. We will use the same approach for all durative actions. 
We introduce the action $start\_flying(L_1,L_2,S)$ to denote that  the plane has started flying from location $L_1$ to location $L_2$ with speed $S$. The variable $S$ ranges over $\{400,600\}$. The remaining actions in the domain are $start\_deplaning(P,L)$, $end\_deplaning(P,L)$, $start\_refueling$, $end\_refueling$, and $end\_flying(L_1,L_2)$.

\noindent
 The  signature consists of the boolean fluents $boarding(P,L)$, $deplaning(P,L)$,  $on\_boa$-$rd(P)$ and $refueling$. We also have non-boolean fluents $location(P)$ and \emph{location(plane)} which range over $\{a,b,c,d,enroute\}$. We have process fluents  \emph{time\_left\_board(P,L)} and  \emph{time\_left\_deplane(P,L)} which are clock functions to count down the time left for person $P$ to board and deplane at location $L$ respectively.  We have process fluent $distance\_$ $left(L_1,L_2)$ to denote the distance that is yet to be covered to reach $L_2$ from $L_1$. Finally, the process fluent $fuel\_level$ denotes the fuel level in the plane's tank. We proceed with defining the effects of each action. The effects of $start\_boarding(P,L)$ are defined using dynamic causal laws
\[ \begin{array}[t]{l}
start\_boarding(P,L)  \mbox{  causes }   boarding(P,L). \\
start\_boarding(P,L) \begin{array}[t]{r} \mbox{  causes  }  time\_left\_board(P,L) = \lambda T. max(0, 30-(T-T_0))  \\  \mbox{ if  } end=T_0. \end{array}
\end{array} \]
The following executability condition says that person P cannot start boarding if he or she is already boarding. 
 \[ \mbox{ impossible } start\_boarding(P,L)  \  if  \  boarding(P,L). \]
The following are causal laws involving $end\_boarding(P,L)$.
\[ \begin{array}[t]{l}
end\_boarding(P,L)  \mbox{ causes } \neg boarding(P,L).\\
end\_boarding(P,L)  \mbox{ causes } on\_board(P). \\
 \mbox{ impossible } end\_boarding(P,L)  \mbox{ if  }  \neg boarding(P,L).
 \end{array}  \]
The following are causal laws involving actions $start\_deplaning(P,L)$ and  \emph{end\_depla ning(P,L)}.
\[ \begin{array}[t]{l}
start\_deplaning(P,L) \mbox{ causes }   deplaning(P,L) \\
start\_deplaning(P,L)  \begin{array}[t]{l} \mbox{  causes  } \\ time\_left\_deplane(P,L)= \lambda T. max(0, 20-(T-T_0))  \\  \mbox{ if  } end=T_0.\end{array} \\
\mbox{impossible } start\_deplaning(P,L)   \mbox{ if }   deplaning(P,L).\\
end\_deplaning(P,L)  \mbox{ causes } \neg deplaning(P,L).\\
end\_deplaning(P,L) \mbox{ causes } \neg on\_board(P).\\
\mbox{impossible } end\_deplaning(P,L) \mbox{  if }  \neg deplaning(P,L).
\end{array}\]
The following are direct effects of refueling. Instead of assuming that it takes 1 hour to fill up from an empty tank we will assume that the rate of refueling is 20 gallons per minute. When we start refueling,  we add fuel to the existing level and the level increases at the rate of 20 gallons per minute. 
\[ \begin{array}[t]{l}
start\_refueling  \begin{array}[t]{l} \mbox{ causes }  fuel\_level=\lambda T. max(750, X + 20*(T-T_0))\\
                                        \mbox{ if } end=T_0, fuel\_level(end) = X.
										 \end{array}\\
start\_refueling \mbox{ causes } refueling. \\
\mbox{ impossible } start\_refueling \mbox{  if  } refueling.\\ 
\mbox{impossible } start\_refueling \mbox{ if } location(plane) =enroute.\\
end\_refueling \mbox{ causes } \neg refueling. \\
\mbox{impossible } end\_refueling \mbox{ if  }location(plane) = enroute.\\
\mbox{ impossible } end\_refueling \mbox{ if } \neg refueling.
\end{array}\]
Next, we define the direct effects of $start\_flying(L_1,L_2,S)$.  
\[ \begin{array}[t]{l}  
start\_flying(L_1,L_2,S)  \mbox{ causes }  location(plane)=enroute.\\
start\_flying(L_1,L_2,S) \begin{array}[t]{l} \mbox{ causes }   distance\_left(L_1,L_2)= \lambda T.  max(0, X - S*(T-T_0)/60)  \\
           \mbox{ if }   distance(L_1,L_2,X), \ end=T_0.  \end{array} \\
start\_flying(L_1,L_2,S)  \begin{array}[t]{l} \mbox{ causes }  fuel\_level= \lambda T. max(0,X -  S*(T-T_0)/ (60*fc(S))) \\  \mbox{ if }   											fuel\_level(end)=X, \ end=T_0 \end{array} \\
\mbox{impossible } start\_flying(L_1,L_2,S) \mbox{ if } location(plane)=enroute.\\
\mbox{impossible } start\_flying(L_1,L_2,S) \mbox{ if } location(plane) \neq L1. \\
\mbox{impossible } start\_flying(L_1,L_2,S) \mbox{ if  } \begin{array}[t]{l}boarding(P,L1), \\ location(P)=L1. \end{array}
\end{array}\]
The following are causal laws involving action $end\_flying(L_1,L_2)$.
\[\begin{array}[t]{l}
end\_flying(L_1,L_2) \mbox{ causes } location(plane)=L_2.\\
\mbox{impossible } end\_flying(L_1,L_2) \mbox{ if  } location(plane) \neq enroute.
\end{array}\]
The following causal laws specify conditions under which each ``ending" action is triggered.
\[ \begin{array}[t]{l}
time\_left\_board(P,L,end) =0   \mbox{  triggers  } end\_boarding(P,L).\\
time\_left\_deplane(P,L, end) =0 \mbox{  triggers  } end\_deplaning(P,L).\\
fuel\_level(end)=750 \mbox{  triggers  } end\_refueling.\\
distance\_left(L_1,L_2,end)=0 \mbox{  triggers  }  end\_flying(L_1,L_2).
\end{array}\] 
The following state constraint encodes common sense knowledge. It states that a person on board a plane is at the same location as the plane.
\[ location(P)=L  \mbox{ if  } location(plane)=L,  on\_board(P).   	\]
The following executability condition states that flying to a location without enough fuel is not allowed. A rational reasoner who wants to reach his or her destination safely will not violate this constraint.
\[\mbox{ impossible }  start\_flying(L_1,L_2,S)  \mbox{ if } \begin{array}[t]{l} fuel\_level(end) = X,\\
                                                               distance(L_1,L_2,Y),\\
                                                               X < Y/fc(S).
\end{array}\]
This concludes the representation of the domain in H.

\section{Translation into logic program}
We solve the planning and scheduling problem by translating the above theory into an Answer Set Prolog program and computing answer sets of the resulting program. Thus, we reduce the problem of finding solutions to the planning and scheduling problem to computing answer sets of the resulting program. 
The translation that was provided in \cite{chi10} is a theoretical translation from statements of H into rules of A-Prolog. However, in this paper we are going to provide a translation that adheres to the syntax of a specific ASP solver. Since we are dealing with continuous functions we chose a solver that, in addition to computing answer sets, is capable of reasoning about constraints over reals. Here is a brief summary of solvers that are available at our disposal.

In the past few years researchers \cite{b09,mgz08} have focused on integrating Answer Set Programming(ASP) and Constraint logic programming(CLP). They came up with new systems that  achieve significant improvement in performance over existing ASP solvers. The following such systems are available: \emph{ACsolver}\cite{mgz08} (and it's successor 
\emph{Luna}\cite{ric10}); \emph{EZCSP} \footnote{http://marcy.cjb.net/ezcsp/index.html}; and \emph{Clingcon}\footnote{http://www.cs.uni-potsdam.de/clingcon/}.   Each solver couples a constraint solver with an answer set solver. However, the coupling varies.  In \emph{Luna} $(ACsolver)$ and $Clingcon$ the coupling is tight whereas in EZCSP the coupling is  loose.  We considered using \emph{Luna} for our purposes but we found out that there are some implementation issues. Besides, the underlying answer set solver is less efficient than the one used by EZCSP.  We also considered using $Clingcon$, however,  the underlying constraint solver only deals with finite domains which is a limitation when dealing with continuous functions. We chose EZCSP as it allows us to reason about constraints over reals.


In EZCSP, the A-Prolog programs are written in such a way that their answer sets encode a constraint satisfaction problem. The system calls the answer set solver,  uses the resulting answer sets to pose a constraint satisfaction problem in the input language of the constraint solver, then calls the constraint solver and combines the solutions returned by both solvers.  The constraint solver can use the results from the answer set solver to solve new constraints but the answer set solver cannot use results from the constraint solver to make new inferences. The current version uses gringo+clasp\footnote{http://potassco.sourceforge.net/} by default as the ASP solver and SICSTUS Prolog as the constraint solver. Both solvers are very fast.


The translation that we present here will be in the input language of EZCSP. Before we begin the translation there are some modeling decisions to be made. These decisions are necessary to overcome some of the implementation issues.  One of the main issues is the implementation of functions. The input language of EZCSP will allow us to represent constants and variables but it is not possible to represent functions of time.  For example, we can translate statements  such as $velocity=5$ directly into the input language of EZCSP but it is not possible to translate a statement such as $height= \lambda T. T^2$ . The reason is that there is no representation for functions in this language.  The alternative is to provide a value for T and obtain the value of the function for that instance. So it is possible to translate the statement $height(5)=\lambda T. T^2(5)=25$.  The constraint solver  may obtain the value for $T$ from other constraints and then use that value to evaluate $height(5)$.

Sorts, actions, boolean fluents and fluents ranging over finite domains will be translated as atoms in the input language of EZCSP. All other fluents ranging over infinite or large domains will be represented as constraint variables. Every constraint involving a  constraint variable will appear in an atom called \emph{required}.  For example, we write $required(height > 5)$ to denote the variable $height$ must be greater than 5.  Process fluents will be represented by constraint variables.  For each process fluent we can use two constraint variables - one to denote the initial value and one to denote the final value in a given state.

Let us look at some statements from our example and see how they are translated into EZCSP rules. Consider the following statement from our example.
\[ start\_refueling  \begin{array}[t]{l} \mbox{ causes }  fuel\_level=\lambda T. max(750, X + 20*(T-T_0))\\
                                        \mbox{ if } end=T_0, fuel\_level(end) = X.
										 \end{array} \]
Since we cannot encode functions in EZCSP we are going to expand the signature to include a boolean fluent called $refueling$, 
a real-valued fluent called $f\_time$ to denote the time at which fueling began and two new real-valued fluents - $f\_initial$ and $f\_final$ to denote the initial and final fuel level in a state. We then replace the above causal law with the following causal laws.
\[  \begin{array}[t]{l} start\_refueling  	\mbox{ causes }   f\_initial = X   \mbox{ if }  f\_final =X.\\  
 start\_refueling  	\mbox{ causes }   f\_time = T   \mbox{ if }  end =T.  \\
start\_refueling  	\mbox{ causes }   refueling.  \\
  f\_final= max( 750, X + 20*(T - T_0))  	\mbox{  if  }   \begin{array}[t]{l} f\_initial = X,
                   										f\_time =T_0, \\
										                  end= T,
										                  refueling.
										                  \end{array} \end{array} \]	
The first two dynamic causal laws say that executing $start\_refueling$ initializes $f\_initial$ to the current fuel level  and $f\_time$ to the end time of the current state. The next dynamic law says that $start\_refueling$ begins the $refueling$ process. Next, we have a state constraint which defines $f\_final$  in terms of $f\_initial$, $f\_time$ and $end$ of a state in which $refueling$ is true.
We consider fluents $f\_initial$, $f\_time$ and $refueling$ to be inertial. These fluents will retain their values until an action causes them to change.  For example, $f\_initial$ will be reset by actions $end\_refueling$, $start\_flying(L_1,L_2,S)$ and $end\_flying(L_1,L_2)$  because each action has a bearing on the fuel level. 	So we will add dynamic causal laws encoding the effects of these actions on $f\_initial$. 	The relationship between the process fluent $fuel\_level$ in the original action description and
$f\_final$ in the modified action description is that for any given state, $f\_final = fuel\_level(end)$.

To translate the above statements into EZCSP we have to first define the various constraint variables. All the real-valued fluents will be translated as constraint variables. 
Since fluents are dependent on state, we have to consider state as one of its parameters during translation. It is important to note that with the assignment of time intervals to states, each state in a transition diagram is unique. 
 Our approach is to assign (non-negative) integers to each state in the order it appears in the trajectory of a transition diagram.
 This is done in the language of EZCSP as follows.
 \[ \begin{array}[t]{l} \#const  \ n=10.\\
      step(0..n). \ \ 
      \# domain \ step(I;I1).
       \end{array} \]
 Here $n$ denotes the length of the trajectory and the $\# domain$ declaration says that variables $I$ and $I1$ range over $[0,n]$ . Next, we will define the various constraint variables parameterized by $step$.
 \[  \begin{array}[t]{l} cspvar(f\_time(I),0,400).  \ \
     cspvar(f\_initial(I),0,750).\\
     cspvar(f\_final(I),0,750). \ \
     cspvar(end(I),0,400). \end{array}
     \]
 The numbers in each declaration specify the range of that variable. Next, the boolean fluent $refueling$ will be translated into the atom $v(X,refueling,I)$ where $X$ ranges over $\{true,false\}$. It says that $X$ is the value of $refueling$ in step $I$. The action $start\_refueling$ is translated as $occurs(start\_refueling,I)$ which says that  $start\_refueling$ occurred in step $I$.
 Finally, here are the EZCSP rules obtained by translating the above causal laws.
 \[  \begin{array}[t]{l}
 required(f\_initial(I1)==f\_final(I)) :- \begin{array}[t]{l} occurs(start\_refueling,I),
 									I1 = I+1. \end{array} \\
required(f\_time(I1)==end(I)) :- \begin{array}[t]{l} occurs(start\_refueling,I), 
                            I1 = I+1. \end{array} \\				
v(true,refueling,I+1) :- occurs(start\_refueling,I).\\ 					
\begin{array}[t]{r}  required(f\_final(I)==max(750,f\_initial(I) + 20*(end(I)-f\_time(I)) )) :-  \\  v(true,refueling,I). \end{array}
\end{array}
 \]     
The rest of the dynamic causal laws are translated using a similar approach. As mentioned before, in EZCSP the answer set solver does not receive any input from the constraint solver to make new inferences. This poses a problem especially when translating triggers. A trigger mentions the conditions under which an action will be executed. It is possible that these  conditions are constraints over time.  Even if these constrains are evaluated, since there is no feedback to the answer set solver, it is not possible to know when an action is triggered.  The author of EZCSP, Marcello Balduccini, has suggested a solution to overcome this problem.  Suppose that we have the following trigger.
\[ fuel\_level(end) = 750 \mbox{  triggers } end\_refueling.\]
Since we have replaced process fluent $fuel\_level$ with other real valued fluents, this trigger is really
\[ f\_initial = X,  f\_time=T,  refueling, end = (750-X)/20 + T    \mbox{ triggers  }  end\_refueling \]
Given the initial fuel level $X$ and the refuel rate of 20 gallons per minute, the expression $(750-X)/20$ gives the number of minutes it takes to fill up the tank. The trigger states that if a state ends at a time when the tank is full then $end\_refueling$ is triggered. Of course, $refueling$ must be also be true in that state. A direct translation of this trigger is
\[
occurs(end\_refueling,I) :- \begin{array}[t]{l}  required(end(I)== (750-f\_initial(I))/20 + f\_time(I)),\\
                            v(true,refueling,I).  \end{array} \] 
This rule will be fired if the body is satisfied.  The atom $v(true,refueling,I)$ is a direct consequence of $start\_refueling$. However, the constraint atom in the body of the rule is not obtained from other rules of the program. Balduccini suggested that it can be generated by adding the rules
\[  \begin{array}[t]{l}1\{p(I), q(I) \}1 :- v(true,refueling,I).  \\ 
          required(end(I) == (750-f\_initial(I))/20 + f\_time(I) ) :- p(I).\\               		             
required(end(I) < (750-f\_initial(I))/20 + f\_time(I) ) :- q(I).     
       \end{array}\]
The choice rule will allow us to generate one of the two constraints.  If no action takes place before the tank fills up then $end\_refueling$ is triggered.  We use the same approach to translate other triggers in the domain. Currently, there is no formal proof stating that this solution will always work. However, the solution is based on answer set programming methodology and EZCSP uses a state-of-the-art answer set solver. We rely on the solver to give us the correct solutions.    										                  						

In our example, the goal is to have Scott and Ernie in city $d$ in less than five and a half hours (330 minutes).  We encode this goal in EZCSP as follows 
\[   goal(I) :-  \begin{array}[t]{l} v(d,location(scott),I), 
            v(d,location(ernie),I),   \\
            required(start(I)<330). \end{array}  \]
where $start(I)$ is a constraint variable denoting the start time of a step. The constraint says that the goal state must be achieved in less than five and a half hours.  It is generated by adding the rules
\[  \begin{array}[t]{l} 1\{g(I),ng(I)\}1.  \\           
                  required(start(I)<330) :- g(I). \\
                  required(start(I) >=330) :- ng(I).
            \end{array} \]
   We add the following rules to say that failure is not an option.
   \[ \begin{array}[t]{l}  success :- goal(I).  \\
:- not \  success. \end{array} \]
 Next, we will talk about the planning component of the program. As in the case of non-continuous domains, we will use Answer Set Programming techniques for generating and testing plans.  A plan is a sequence of actions. We will be generating all possible sequences of actions that will lead us to our goal. In the process of generating these plans we use the constraints in our program to test these plans and discard the ones that violate the constraints.  A simple way to generate plans is to use the choice rule
 \[ \{occurs(A,I): action(A)\} :- step(I), I<n. \]
 where $n$ is the length of the plan. Note that we are going to generate only the initiating actions for example, $start\_refueling$ etc. This is because all the terminating actions have triggers associated with them and there is no need to generate them again.  Since we are dealing with continuous time, in addition to  computing the sequences of actions, we will also determine when these actions will take place. This is the reason why planning in hybrid domains also involves scheduling.  According to the problem specification, we have a constraint that specifies the time within which the goal has to be achieved i.e. five and a half hours. We can specify this constraint in EZCSP as follows.
\[  required(end(I)< 330 ) :- occurs(A,I), action(A). \]
 The rules states that all action occurrences must take place before five and a half hours.  This constraint gives a very broad range of scheduling possibilities. The times during which actions will take place could now range over intervals of time. For example, in order to achieve our goal  it may be required that Scott has to board the plane within the first 50 minutes and that the plane has to depart city $c$ no later than three hours and 45 minutes into the trip.  Dealing with time intervals is an issue because it does not allow us to compute values of fluents at specific time points.  More issues arise when fluents are defined in terms of other fluents whose values are unknown. To overcome these issues we decided to assign specific times to the initiating actions. The times for the terminating actions were determined by our triggers. Here is a solution given by EZCSP to our planning and scheduling problem.
\[  \begin{array}[t]{ll}  occurs(start\_boarding(scott,a),0)  & end(0) =5.0\\
                                    occurs(end\_boarding(scott,a),1)   & end(1)=35.0\\
                                    occurs(start\_flying(a,c,400),2)  & end(2)=40.0\\
                                    occurs(end\_flying(a,c),3)  & end(3)=190.0 \\
                                    occurs(start\_refueling,4) & end(4)=195.0\\ 
                                    occurs(start\_boarding(ernie,c),5)  & end(5)=197.0\\
                                    occurs(end\_refueling,6)   & end(6)= 224.16 \\
                                    occurs(end\_boarding(ernie,c),7) & end(7)= 227.0 \\
                                    occurs(start\_flying(c,d,600),8)  & end(8)=229.0 \\
                                    occurs(end\_flying(c,d),9)  & end(9)=329.0
      \end{array}\]
It is necessary to refuel in city $c$ because there is not enough fuel to travel to city $d$.  
A shorter plan can be obtained by allowing concurrent actions. For example, refueling and boarding can start at the same time. 
The running times and  the performance of the solver will be discussed in a longer version of the paper.

\section{Conclusions and Future Work}
In this paper, we presented an Action Language-Answer Set Programming based approach to solving planning and scheduling problems in hybrid domains. We used action language H to model a planning and scheduling example and translated the resulting theory into an A-Prolog program. We used a hybrid solver called EZCSP to compute the answer sets of the resulting program.  Our approach overcomes the limitations of existing formalisms such as $PDDL+$.
We believe that our approach can be applied to any planning and scheduling problem in hybrid domains.

In the future, we would like to model several benchmark examples and compare the performance of  EZCSP with existing planners. Some of the planners used for industry-sized problems are domain-specific \cite{bj04}. It will be useful to investigate why some of these planners work really well.  There are areas for improvement including the efficiency of solvers, the expressiveness of the input language and so on.


\bibliographystyle{plain}
\bibliography{SanWatFestFinal}

  \end{document}